\documentclass[lettersize,journal]{IEEEtran}
\usepackage{amsmath,amsfonts}
\usepackage{algorithmic}
\usepackage{algorithm}
\usepackage{array}
\usepackage[caption=false,font=normalsize,labelfont=sf,textfont=sf]{subfig}
\usepackage{textcomp}
\usepackage{stfloats}
\usepackage{url}
\usepackage{verbatim}
\usepackage{graphicx}
\usepackage{cite}
\usepackage{multirow}
\usepackage{siunitx}
\begin{document}

\title{Autoencoder-Based Denoising of Muscle Artifacts in ECG to Preserve Skin Nerve Activity (SKNA) for Cognitive Stress Detection}

\author{Farnoush Baghestani, \IEEEmembership{Student Member, IEEE}, Jihye Moon, \IEEEmembership{Student Member, IEEE}, Youngsun Kong, \IEEEmembership{Member, IEEE}, Ki Chon, \IEEEmembership{Fellow, IEEE}
\thanks{Farnoush Baghestani, Jihye Moon, Youngsun Kong, and Ki Chon are with the department of Biomedical Engineering, University of Connecticut, Storrs, CT 06269 USA.}}



\maketitle

\begin{abstract}
The sympathetic nervous system (SNS) plays a central role in regulating the body’s responses to stress and maintaining physiological stability. Its dysregulation is associated with a wide range of conditions, from cardiovascular disease to anxiety disorders. Skin nerve activity (SKNA) extracted from high-frequency electrocardiogram (ECG) recordings provides a noninvasive window into SNS dynamics, but its measurement is highly susceptible to electromyographic (EMG) contamination. Traditional preprocessing based on bandpass filtering within a fixed range (e.g., 500–1000 Hz) is susceptible to overlapping EMG and SKNA spectral components, especially during sustained muscle activity. We present a denoising approach using a lightweight one-dimensional convolutional autoencoder with a long short-term memory (LSTM) bottleneck to reconstruct clean SKNA from EMG-contaminated recordings. Using clean ECG-derived SKNA data from cognitive stress experiments and EMG noise from chaotic muscle stimulation recordings, we simulated contamination at realistic noise levels (–4 dB, –8 dB signal-to-noise ratio) and trained the model in the leave-one-subject-out cross-validation framework. The method improved signal-to-noise ratio by up to 9.65 dB, increased cross correlation with clean SKNA from 0.40 to 0.72, and restored burst-based SKNA features to near-clean discriminability (AUROC $\geq$ 0.96). Classification of baseline versus sympathetic stimulation (cognitive stress) conditions reached accuracies of 91–98\% across severe noise levels, comparable to clean data. These results demonstrate that deep learning–based reconstruction can preserve physiologically relevant sympathetic bursts during substantial EMG interference, enabling more robust SKNA monitoring in naturalistic, movement-rich environments.
\end{abstract}


\section{Introduction}
\IEEEPARstart{A}{utonomic} regulation is the body’s ability to adjust heart rate, blood pressure, and other vital functions to meet the demands of physical and psychological stress. This regulation depends on two complementary branches: the sympathetic branch, which accelerates cardiovascular activity, and the parasympathetic branch, which promotes recovery and conservation of energy. The quality of this balance strongly influences overall health. Persistent disruption, particularly sympathetic overdrive, is implicated in hypertension, heart failure, arrhythmia, and anxiety disorders. Quantifying sympathetic nervous system (SNS) activity can therefore provide critical insight into both health and disease \cite{sinski2006study}, \cite{hering2015role}.

ECG-derived Skin nerve activity (SKNA) has recently emerged as a promising alternative to the invasive microneurography for assessing SNS dynamics. SKNA refers to the electrical signals of sympathetic nerves recorded from the skin’s surface, typically obtained by high-frequency sampling of ECG electrodes (often $\geq$ 2 kHz, sometimes termed NeuECG) \cite{kusayama2020simultaneous}. SKNA has been observed to precede both the onset and termination of atrial tachycardia and fibrillation \cite{uradu2017skin}. Elevated resting SKNA has been associated with a greater risk of subsequent ventricular arrhythmia episodes. \cite{zhang2019characterization}. Beyond arrhythmia research, SKNA has shown potential in orthostatic intolerance, where symptomatic episodes are frequently preceded by SKNA bursts even in the absence of tachycardia \cite{lee2022skin}. In obstructive sleep apnea, large SKNA bursts accompany arousals from apnea, particularly during NREM2 sleep \cite{he2020skin}. In patients in intensive care, altered temporal fluctuations of SKNA have been shown to predict both short- and long-term mortality \cite{chen2021complex}.

Because SKNA can be measured noninvasively, assessing SNS activation through SKNA holds promise for practical, real-world applications. These include portable cardiovascular disease management, emotion recognition, cognitive fatigue monitoring, pain assessment, and detection of sleep deprivation.  For example SKNA has been explored as a biomarker of physical fitness, indicating individual differences in sympathetic tone and the capacity to adapt to physiological stressors \cite{liu2021skin}. A portable device for SKNA recording has been developed in a recent study \cite{xing2022design}. However, our prior study has shown that the quality of SKNA-based SNS measurements is highly affected by electromyographic (EMG) contamination \cite{baghestani2024towards}. Removing this muscle noise is therefore critical for preserving the integrity of SKNA signals.

Previous studies have largely relied on traditional preprocessing methods, most commonly fixed-range bandpass filtering, which can fail when SKNA and EMG spectral components overlap. In standard preprocessing, SKNA is extracted by isolating high-frequency components of ECG recordings through bandpass filtering, commonly in the 500–1000 Hz range. This step is intended to suppress lower-frequency cardiac, muscle, and motion-related artifacts, as well as higher-frequency noise \cite{kusayama2020simultaneous}. The filtered SKNA is full wave rectified and integrated over a short moving window (e.g., 100 ms) to produce the integrated SKNA (iSKNA) trace, which smooths burst patterns and facilitates detection of sustained changes in sympathetic tone. While this procedure is widely used, it is not optimal. Some studies have attempted to extract nerve activity with higher precision, pronouncing physiologically relevant activity \cite{kong2025new}, \cite{su2025sympathetic}. However, aside from the initial bandpass filtering step, there has been no systematic attempt to remove muscle noise from SKNA recordings. In fact, in our previous work we were the first to explicitly identify EMG contamination from skeletal muscles as a critical limitation for SKNA measurement \cite{baghestani2024towards}. EMG power is strongly present even within the high-frequency range typically used for SKNA extraction. Consequently, movement and muscle activity can mask SKNA bursts with much larger EMG spikes, severely degrading the fidelity of the measured sympathetic activity. This problem hampers continuous monitoring of SKNA in realistic scenarios where perfectly still conditions cannot be guaranteed. Recognizing this issue, in our preliminary study, we introduced an approach for detecting and consequently discarding muscle noise affected segments of the recordings \cite{baghestani2024towards}. This early effort demonstrated the feasibility of algorithmically distinguishing genuine SKNA from noise. Nonetheless, an effective method for removing EMG artifacts from SKNA signals and reconstructing them remains an unexplored area.

In parallel, the biosignals research community has increasingly used deep learning (DL) to address similar denoising problems. Modern deep learning models, such as convolutional neural networks and autoencoders, have shown demonstrable success in separating physiological signals from noise in ECG, PPG, and EEG recordings \cite{dasan2021novel}, \cite{wang2022ecg}, \cite{lin2023ecg}, \cite{nejad2024enhancing}, \cite{mohagheghian2023noise}. Rather than relying on fixed filters, these models, using various filters based on the patterns in data, learn to recognize the complex patterns of the true signal from artifacts. Such approaches can preserve clinically relevant information that might be distorted or lost by conventional filtering.

Motivated by these advances, our work applies a DL solution to SKNA denoising. Specifically, we propose a lightweight 1-D convolutional neural network with recurrent (LSTM) layers as an autoencoder to reconstruct a clean SKNA signal from an EMG-contaminated data. To our knowledge, this represents the first application of DL for denoising EMG-contaminated SKNA measurements. The goal is to reliably preserve sympathetic nerve discharges even during sustained muscle movement, enabling continuous and accurate SNS monitoring in real-world conditions. In the following sections, we detail the methodology of the proposed approach and demonstrate its effectiveness in isolating SKNA from EMG artifacts.

\section{DATASET}
\subsection{SKNA Data Collection}
This study involves a dataset originally collected for a preliminary study examining the effect of muscle noise interference on SKNA signals’ integrity \cite{baghestani2024towards}. The study involved 12 healthy adults (7 females, 5 males; aged 25–32). All participants provided informed consent, and the study protocol was approved by the Institutional Review Board at the University of Connecticut.

ECG data were acquired using a standard two-lead configuration (leads II and III), with electrodes placed on the wrists and ankles, and a ground electrode positioned below the right rib cage. Recordings were made in a supine position using a Bio Amp connected to a PowerLab system (Sydney, Australia), sampled at 10 kHz via LabChart Pro 7 software.

In the dataset, one of the two ECG channels was subjected to voluntary muscle movements, while the other remained stationary to minimize artifacts. In this study, we use a subset of the original data: specifically, ECG recordings from the channel that remained free of movement artifacts (“clean” channel) during the following four phases: a 2-minute resting period (Baseline recording 1), a 5-minute SNS stimulation through Stroop cognitive task, a second 2-minute resting period, and a second 5-minute Stroop cognitive task.

The Stroop task \cite{scarpina2017stroop} is a widely used paradigm in cognitive neuroscience for inducing and measuring cognitive load. It exploits the conflict between automatic processes (reading a word) and controlled processes (naming the ink color), requiring participants to suppress the more dominant reading response. This interference reliably engages executive control mechanisms and activates stress-related neural pathways, making it an effective tool for studying attention, processing speed, and cognitive stress.

During the Stroop tasks, participants were presented with color–word stimuli on a tablet screen, where the text color often conflicted with the word itself (e.g., the word blue displayed in red). Participants were instructed to name the ink color aloud, a process designed to induce cognitive stress.

\subsection{EMG Dataset}
To simulate muscle artifact in our analysis, we used EMG signals from an independent dataset consisting of biceps muscle recordings under controlled chaotic stimulation \cite{khodadadi2023dataset}. This dataset was developed using a method where the Rossler chaotic equation was employed to generate stimulation signals applied to the musculocutaneous nerve. The resulting chaotic EMG signals were recorded from the biceps brachii muscle of 10 healthy right-handed individuals (6 males, 4 females; average age: 31.2 years).

Each participant underwent two recording sessions, yielding 20 raw EMG signals in total. Signals were captured using surface electrodes and a PowerLab 26T system, sampled at 4000 Hz for a duration of 414 seconds per recording. The stimulation signal, derived from the z-component of the Rossler system, was normalized and applied using a National Instruments device (BNC-2090), with parameter $\alpha$ varied from 0.1 to 0.3 to induce chaotic activity. These EMG signals, rich in nonlinear and chaotic characteristics, were used in this study as additive noise to evaluate the robustness of our denoising method under realistic muscle artifact conditions. We verified via spectral analysis that EMG noise was broadly present across the 0–2000 Hz band.

Two noise levels were simulated to evaluate the robustness of the proposed reconstruction method: –4 dB and –8 dB SNR. The lower SNR level was included because our prior study showed that EMG contamination in SKNA recordings can reach –8 dB under conditions of strong muscle activation, making it a realistic worst-case scenario for signal recovery \cite{baghestani2024towards}. The procedure for adding muscle noise to the ECG data is described in detail in Section \ref{subsec:training}.

\subsection{Data Preprocessing and Segmentation}
In this study, MATLAB was used to read the “.adicht” files generated by LabChart software. SKNA signals were extracted by applying a bandpass filter to the ECG recordings in the 500–1000 Hz range. A notch filter was then applied at 762 Hz to eliminate device-specific noise. The filtered signals were resampled from 10 kHz to 2048 Hz and segmented into 1-second windows. Each segment was labeled according to its corresponding condition, either baseline or Stroop task. The processed data for each subject was stored in HDF5 format for subsequent analysis in Python. The same preprocessing pipeline, bandpass and notch filtering, followed by resampling and segmentation, was applied to the muscle noise (MN) signals.

\section{METHODOLOGY}
The methodology integrates signal reconstruction, feature extraction, and classification to evaluate the utility of denoised SKNA for detecting sympathetic activation induced by cognitive stress. First, we introduce a lightweight autoencoder designed to remove EMG noise while preserving temporal patterns of SKNA. Its performance is benchmarked against traditional bandpass filtering (BPF) to demonstrate the limitations of filtering alone and underscore the need for more advanced approaches such as deep learning-based filtering. Model training and evaluation follow a leave-one-subject-out cross-validation strategy to ensure generalizability. After denoising the SKNA, features relevant to sympathetic nerve activity are extracted and assessed for their preservation under both baseline and cognitive stress conditions. Statistical analyses quantify whether reconstructed signals maintain sensitivity to condition-related changes, while classification experiments test their ability to distinguish stress states. Together, these steps establish whether the proposed denoising approach enables more reliable measurement of sympathetic activity than BPF alone.

\begin{figure*}
    \centering
    \includegraphics[width=0.98\linewidth]{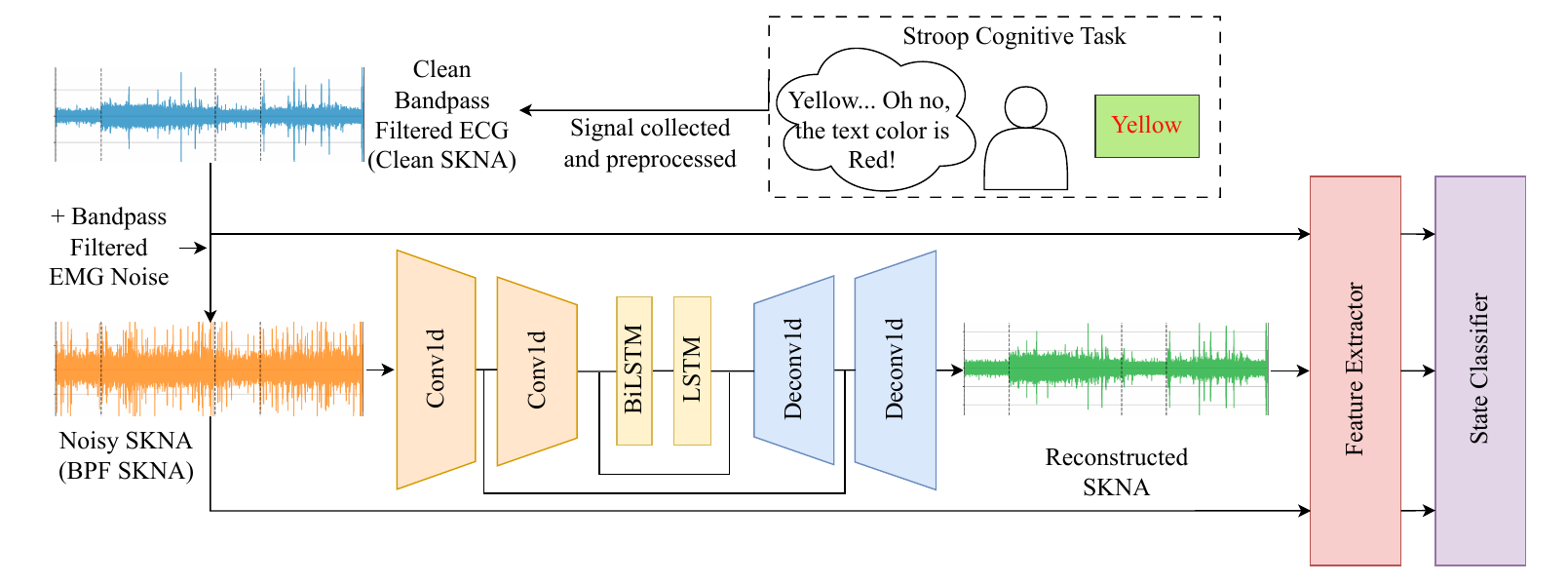}
    \caption{An Overview of The Methodology}
    \label{fig:teaser}
\end{figure*}

\subsection{Proposed Autoencoder}
\subsubsection{Model Architecture}
In this work, we propose an autoencoder to denoise SKNA signals. The proposed model is a lightweight 1-D convolutional autoencoder with an LSTM bottleneck, as shown in figure \ref{fig:teaser} The encoder comprises two convolutional blocks. Each block applies a 1-D convolution (kernel size = 3, stride = 2, padding = 1), followed by batch normalization, dropout (0.2), and ReLU activation. The first convolution block maps the input from one channel to 16 feature maps, and the second convolution block increases the representation to 32 feature maps. Intermediate outputs from both convolutional layers are stored as residual connections for later use in the decoder.

The encoded sequence is passed through two stacked recurrent layers. The first is a bidirectional LSTM with a hidden size of 32, producing a 64-dimensional feature vector at each time step. The second is a unidirectional LSTM with a hidden size of 32, reducing the sequence back to 32 features per step. This bottleneck captures temporal dependencies.

The decoder mirrors the encoder structure using transposed convolutional layers. The first deconvolutional block down samples the sequence from 32 to 16 channels (kernel size = 3, stride = 2, padding = 1, output padding = 1), with batch normalization, dropout (0.1), and ReLU activation. Before this step, the bottleneck output is element-wise added to the second encoder residual. The second transposed convolution maps 16 channels back to a single output channel (same kernel/stride/padding/output padding configuration), again preceded by residual addition from the first encoder layer. Residual connections between corresponding encoder and decoder stages help preserve fine-grained temporal details and stabilize training.

In summary, the forward pass consists of encoding the input through two convolutional stages, processing the resulting sequence in the LSTM bottleneck, and reconstructing the signal through two deconvolutional stages, with skip connections integrated at each level.

The proposed denoising network was designed to balance reconstruction accuracy with computational efficiency, enabling future deployment in portable or real-time SKNA monitoring systems. SKNA recordings have strong temporal structure and relatively low spatial complexity, making 1-D convolutions well-suited for extracting local waveform features while keeping the parameter count low. The shallow encoder–decoder design minimizes memory and computation costs, which is critical for processing high-frequency ECG-derived signals on resource-constrained hardware.

The LSTM bottleneck was included to capture long-range temporal dependencies that pure convolutional layers may miss, such as the temporal patterning of sympathetic bursts. Using a bidirectional LSTM followed by a unidirectional LSTM allows the network to incorporate both past and future context during training, while producing a causal output suitable for real-time inference.

Residual connections between corresponding encoder and decoder layers help preserve fine-grained temporal details that are easily lost during aggressive down sampling, and they stabilize gradient flow during training. The overall architecture therefore provides sufficient capacity to distinguish SKNA from overlapping EMG noise without overfitting, while retaining computational efficiency for fast inference and potential integration into wearable or bedside devices.

\subsubsection{Autoencoder Model Training}
\label{subsec:training}
To ensure class balance within each training batch, we used a custom batch sampler that enforces equal representation from each class. The class-aware batch sampler, groups samples based on their labels and constructs each batch by randomly selecting an equal number of examples from each class. A batch size of 32 was used. For evaluation, a standard data loader without shuffling was employed to preserve the original order of the test segments.

The network was trained for 200 epochs using mean squared error (MSE) loss and the Adam optimizer, with an initial learning rate of 0.001, no weight decay, and default parameters otherwise.

To address data imbalance, a time-shifted version of the MN data was prepared for potential use for data augmentation. For each subject, training and test sets were assigned once and kept fixed throughout the experiments. To add the muscle noise segments to the ECG signals, for each test subject, a random subject from the MN dataset was selected and used exclusively for adding noise to the test set, while the remaining MN recordings were used for training. MN segments were randomly selected and added to the clean signal segments. Due to label imbalance between conditions, we augmented the training data with additional noisy baseline segments using the shifted version of the MN signals. Finally, the signal-to-noise ratio (SNR) was computed across all segments, and a single global scaling factor was applied to the MN signals to match a desired SNR level.

Both the training and test data were then normalized using the mean and standard deviation computed from the training set. We also created additional segments with 50\% overlap, which helps to avoid sharp transitions when reconstructing the signal.

\subsection{Autoencoder Model Performance Evaluation}
To ensure generalizability, a leave-one-subject-out (LOSO) cross-validation evaluation strategy was employed. In each fold, the model was trained on data from all but one subject and tested on the remaining subject. This process was repeated until every subject had served as the test subject exactly once. Performance metrics were then averaged across folds to provide an overall evaluation of the model’s generalization ability to unseen subjects.

After reconstructing the denoised signals from the overlapping segments, quantitative performance was evaluated using four metrics computed between the denoised output and the corresponding clean reference: signal-to-noise ratio (SNR), mean squared error (MSE), mean absolute error (MAE), and cross-correlation.

\subsection{Feature Extraction from SKNA}
\begin{figure*}
    \centering
    \includegraphics[width=0.98\linewidth]{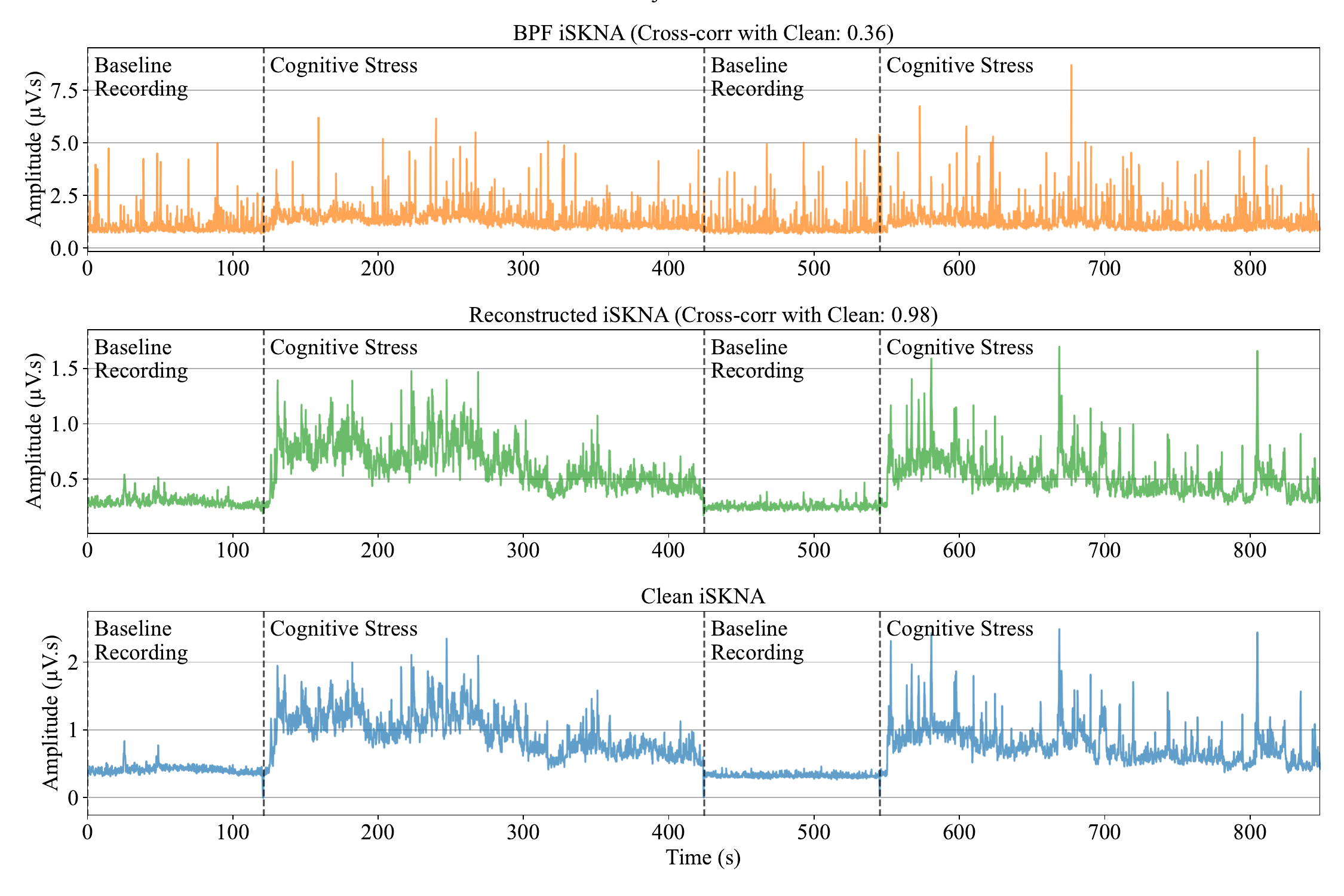}
    \caption{Example iSKNA traces for BPF, reconstructed, and clean signals during baseline and stimulation periods at SNR –4 dB. The top panel shows the EMG-contaminated (BPF) iSKNA with reduced correlation to the clean reference (r = 0.36). The middle panel shows the reconstructed iSKNA, in which burst patterns and baseline levels closely match the clean reference (r = 0.98). The bottom panel shows the clean iSKNA. Vertical dashed lines indicate transitions between baseline recording and SNS stimulation through Stroop cognitive task periods.}
    \label{fig:iskna_example}
\end{figure*}

\begin{figure*}
    \centering
    \includegraphics[width=0.98\linewidth]{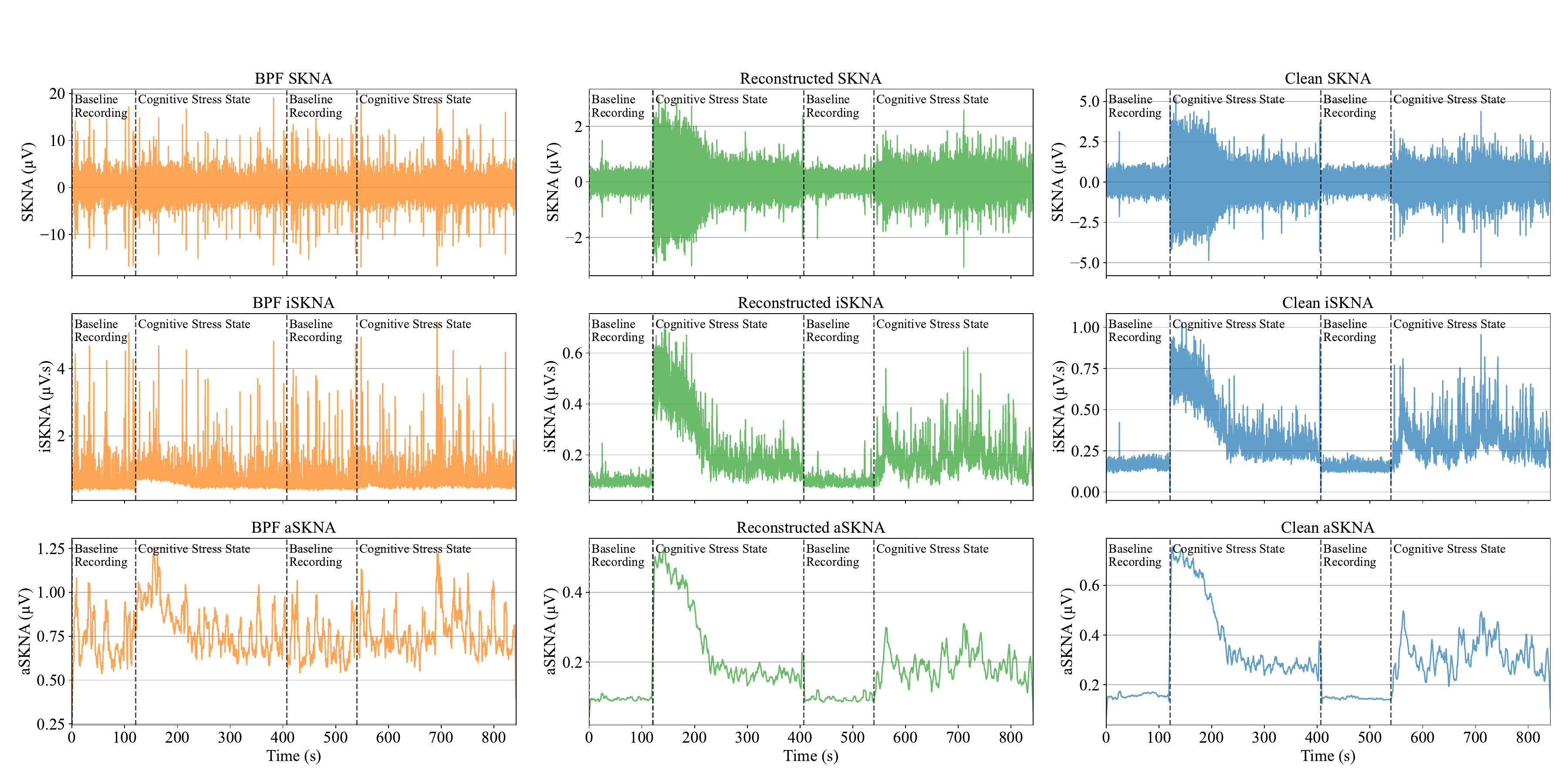}
    \caption{Example SKNA, iSKNA, and aSKNA traces for BPF, reconstructed, and clean signals from a representative subject under simulated EMG contamination at SNR –8 dB. Each row shows the same signal type across baseline and SNS stimulation through Stroop cognitive task periods. Reconstruction markedly improves similarity to the clean reference, with recovery of burst patterns and amplitude profiles despite the severe noise level. Vertical dashed lines indicate transitions between baseline and SNS stimulation through Stroop cognitive task periods.}
    \label{fig:all3_example}
\end{figure*}

Based on common practice in microneurography and SKNA signal processing, we extracted additional derived signals to facilitate clearer visualization of sympathetic bursts and support future feature extraction. Following signal reconstruction, the SKNA signals were rectified by taking the absolute value of the denoised output. The integral of the rectified SKNA signal, termed iSKNA, was computed using a leaky integrator function with time constant $\tau$ = 0.1. To obtain the average SKNA (aSKNA), a moving average filter was applied to the iSKNA signal over a 5-second window. Figures \ref{fig:iskna_example} and \ref{fig:all3_example} illustrate examples of reconstructed SKNA and the corresponding derived iSKNA and aSKNA signals.

To characterize sympathetic nerve activity from the processed SKNA signals, we computed statistical and burst-related features from iSKNA. Signals were optionally divided into non-overlapping windows of length 10 seconds. For each window, both the mean and standard deviation (std) of the signals were calculated. These statistics reflect the central tendency and variability of sympathetic nerve activity within each interval.

Bursts were identified as periods where the iSKNA signal exceeded a threshold defined per subject and signal type (BPF, reconstructed, clean). The threshold was computed as the mean plus three times the standard deviation of the iSKNA values from baseline (group0) segments before windowing. Within each window, the following burst features were extracted:

\begin{itemize}
\item{Burst Count: Number of bursts per minute.}
\item{Burst Duration: Percentage of window time spent above the threshold.}
\item{Burst Amplitude: Average peak amplitude of detected bursts.}
\item{Burst Total Area: Total area under the bursts above threshold, normalized to $\mu$V·min.}
\end{itemize}
Including the mean of iSKNA and standard deviation of iSKNA, we used all features for analysis.

\subsection{Statistical Analysis for Feature Preservation Assessment}
The statistical analysis aimed to evaluate how well each extracted feature reflects sympathetic activation and whether it is preserved after reconstruction in the presence of noise. For each feature, an omnibus Kruskal–Wallis test was first performed across all six experimental groups, defined by the combination of three signal types (BPF, clean, and reconstructed) and two conditions (baseline and stimulation). This non-parametric test assessed whether at least one group differed significantly from the others, without assuming normality of the feature distributions. P-values below 0.05 were considered statistically significant.

If the omnibus test indicated a potential difference, post-hoc pairwise analyses were conducted using the Wilcoxon signed-rank test for paired samples. Comparisons were restricted to two sets of contrasts of primary interest:
\begin{enumerate}
    \item Baseline vs. stimulation within each signal type, to evaluate condition-related changes.
    \item Clean signals compared to BPF and reconstructed signals within the same condition, to assess the effect of noise and reconstruction on feature preservation.
\end{enumerate}

For the baseline vs. stimulation contrasts within each signal type, two additional effect size and discriminability measures were calculated:
\begin{itemize}
    \item Fisher’s ratio, defined as the squared difference between group means divided by the sum of group variances, quantifying the separability between the two conditions.
    \item Area Under the Receiver Operating Characteristic Curve (AUROC), treating baseline and stimulation as binary classes, with values closer to 1 indicating stronger discriminative capability of the feature.
\end{itemize}

\subsection{Cognitive Stress Detection}
The goal of the classification analysis was to assess how well features derived from each signal type (BPF, clean, reconstructed) could distinguish between baseline and cognitive stress conditions. This evaluation reflects the practical utility of the reconstructed SKNA signals for detecting sympathetic activation associated with cognitive load. Features were standardized prior to model training.

Three classifiers were tested: Random Forest (100 trees, default parameters), Support Vector Machine (RBF kernel, probability estimates enabled), and Logistic Regression (maximum 1000 iterations). The classification subjects were the same participants whose data were used for training and testing the autoencoder; however, classification performance was assessed using LOSO cross-validation, ensuring subject independence between training and test sets. For each fold, accuracy, area under the receiver operating characteristic curve (AUC), sensitivity, specificity, and F1 score were computed.

\section{RESULTS AND DISCUSSION}
The proposed autoencoder was trained and evaluated using only the raw SKNA signal as input. Neither iSKNA nor aSKNA were provided to the network; instead, these derived signals were computed from the reconstructed SKNA output for downstream analysis and visualization. The iSKNA and aSKNA traces were used to assess whether the reconstruction preserved physiologically relevant sympathetic burst structure and to facilitate clearer interpretation in figures. aSKNA was not used in any quantitative evaluation and appears only in illustrative plots for visual context.

Two noise levels were simulated to evaluate the robustness of the proposed reconstruction method: –4 dB and –8 dB SNR. The lower SNR level was included because prior studies have reported that EMG contamination in SKNA recordings can reach –8 dB under conditions of strong muscle activation, making it a realistic worst-case scenario for signal recovery \cite{baghestani2024towards}.

\subsection{Reconstruction}
The proposed autoencoder improved SKNA reconstruction quality under both simulated noise levels (SNR –4 dB and –8 dB), as reflected in all reported metrics (Tables \ref{tab:performance-4dB} and \ref{tab:performance-8dB}).

At SNR –4 dB, denoised SKNA increased cross-correlation coefficient with the clean reference from 0.53 (BPF) to 0.72 overall, with a corresponding improvement in iSKNA correlation from 0.32 to 0.88. Pointwise error was markedly reduced: MSE decreased from 0.91 to 0.18, MAE from 0.60 to 0.31. The reconstructed SKNA also showed a reversal of noise-induced degradation in SNR, improving from –4.00 dB to 3.06 dB.

Under the more severe SNR –8 dB condition, reconstruction again substantially improved fidelity. SKNA correlation increased from 0.40 (BPF) to 0.68, and iSKNA correlation from 0.15 to 0.72. MSE decreased from 1.07 to 0.12, MAE from 0.65 to 0.25, and SNR improved from –7.18 dB to 2.47 dB.

These results demonstrate that the model effectively suppressed broadband EMG contamination while preserving the temporal structure of SKNA and its integrated form. Although reconstruction performance was modestly reduced at –8 dB compared to –4 dB, substantial SNR gains, large correlation increases, and error reductions were consistently achieved across both baseline and stimulation conditions.

\begin{table}[!t]
\caption{Reconstruction performance for SKNA and iSKNA under simulated EMG contamination at SNR~\SI{-4}{\decibel}. Values are mean ± standard deviation, with 95\% confidence intervals. BPF: Bandpass filtered, Recon: Reconstructed output of the autoencoder.
\label{tab:performance-4dB}}
\centering
\begin{tabular}{|c|c|c|c|c|}
\hline Metric & Signal & Baseline & Stimulation & Overall \\
\hline Corr & BPF & $0.38 \pm 0.09$ & $0.57 \pm 0.01$ & $0.53 \pm 0.00$\\
 & & [0.33, 0.43] & [0.56, 0.57] & [0.53, 0.53]\\
\hline Corr & Recon & $0.60 \pm 0.17$ & $0.73 \pm 0.11$ & $0.72 \pm 0.11$ \\
 & & [0.50, 0.69] & [0.67, 0.79] & [0.65, 0.78] \\
\hline Corr & BPF & $0.07 \pm 0.03$ & $0.30 \pm 0.10$ & $0.32 \pm 0.10$ \\
 & (iSKNA) & [0.05, 0.09] & [0.24, 0.36] & [0.26, 0.37] \\
\hline Corr & Recon & $0.44 \pm 0.21$ & $0.85 \pm 0.09$ & $0.88 \pm 0.09$ \\
 & (iSKNA) & [0.32, 0.56] & [0.79, 0.90] & [0.83, 0.93] \\
\hline MSE & BPF & $0.90 \pm 0.62$ & $0.91 \pm 0.65$ & $0.91 \pm 0.64$ \\
 &  & [0.55, 1.25] & [0.54, 1.28] & [0.55, 1.27] \\
\hline MSE & Recon & $0.11 \pm 0.07$ & $0.21 \pm 0.14$ & $0.18 \pm 0.12$ \\
 &  & [0.07, 0.15] & [0.13, 0.29] & [0.12, 0.25] \\
\hline MAE & BPF & $0.60 \pm 0.21$ & $0.60 \pm 0.21$ & $0.60 \pm 0.21$ \\
 &  & [0.48, 0.72] & [0.48, 0.72] & [0.48, 0.72] \\
\hline MAE & Recon & $0.25 \pm 0.09$ & $0.34 \pm 0.12$ & $0.31 \pm 0.11$ \\
 &  & [0.20, 0.30] & [0.27, 0.41] & [0.25, 0.38] \\
\hline SNR & BPF & $-7.83 \pm 2.40$ & $-3.26 \pm 0.32$ & $-4.00 \pm 0.00$ \\
 &  & [-9.19, -6.47] & [-3.44, -3.08] & [-4.00, -4.00] \\
\hline SNR & Recon & $1.27 \pm 2.74$ & $3.23 \pm 1.48$ & $3.06 \pm 1.51$ \\
 &  & [-0.28, 2.82] & [2.40, 4.07] & [2.21, 3.92] \\
\hline
\end{tabular}
\end{table}

\begin{table}[!t]
\caption{Reconstruction performance for SKNA and iSKNA under simulated EMG contamination at SNR~\SI{-8}{\decibel}. Values are mean ± standard deviation, with 95\% confidence intervals. BPF: Bandpass filtered, Recon: Reconstructed output of the autoencoder.
\label{tab:performance-8dB}}
\centering
\begin{tabular}{|c|c|c|c|c|}
\hline Metric & Signal & Baseline & Stimulation & Overall \\
\hline Corr & BPF & $0.37 \pm 0.00$ & $0.25 \pm 0.07$ & $0.40 \pm 0.02$ \\
 & & [0.37, 0.37] & [0.21, 0.29] & [0.39, 0.41] \\
\hline Corr & Recon & $0.67 \pm 0.10$ & $0.56 \pm 0.15$ & $0.68 \pm 0.10$ \\
 & & [0.62, 0.73] & [0.47, 0.65] & [0.62, 0.74] \\
\hline Corr & BPF & $0.16 \pm 0.06$ & $0.04 \pm 0.03$ & $0.15 \pm 0.05$ \\
 & (iSKNA) & [0.12, 0.19] & [0.02, 0.05] & [0.13, 0.18] \\
\hline Corr & Recon & $0.78 \pm 0.14$ & $0.35 \pm 0.18$ & $0.72 \pm 0.14$ \\
 & (iSKNA) & [0.71, 0.86] & [0.24, 0.45] & [0.64, 0.81] \\
\hline MSE & BPF & $1.09 \pm 0.77$ & $1.13 \pm 0.78$ & $1.07 \pm 0.77$ \\
 &  & [0.66, 1.53] & [0.69, 1.57] & [0.64, 1.51] \\
\hline MSE & Recon & $0.10 \pm 0.07$ & $0.06 \pm 0.04$ & $0.12 \pm 0.08$ \\
 &  & [0.06, 0.14] & [0.04, 0.08] & [0.07, 0.16] \\
\hline MAE & BPF & $0.65 \pm 0.25$ & $0.65 \pm 0.25$ & $0.65 \pm 0.25$ \\
 &  & [0.51, 0.79] & [0.52, 0.79] & [0.51, 0.79] \\
\hline MAE & Recon & $0.23 \pm 0.08$ & $0.18 \pm 0.06$ & $0.25 \pm 0.09$ \\
 &  & [0.19, 0.27] & [0.14, 0.21] & [0.20, 0.30] \\
\hline SNR & BPF & $-8.00 \pm 0.00$ & $-12.02 \pm 2.54$ & $-7.18 \pm 0.43$ \\
 &  & [-8.00, -8.00] & [-13.45, -10.58] & [-7.42, -6.94] \\
\hline SNR & Recon & $2.38 \pm 1.03$ & $0.97 \pm 2.03$ & $2.47 \pm 1.07$ \\
 &  & [1.80, 2.96] & [-0.18, 2.12] & [1.87, 3.07] \\
\hline
\end{tabular}
\end{table}

\subsection{Feature Preservation}

\begin{figure*}
    \centering
    \includegraphics[width=0.98\linewidth]{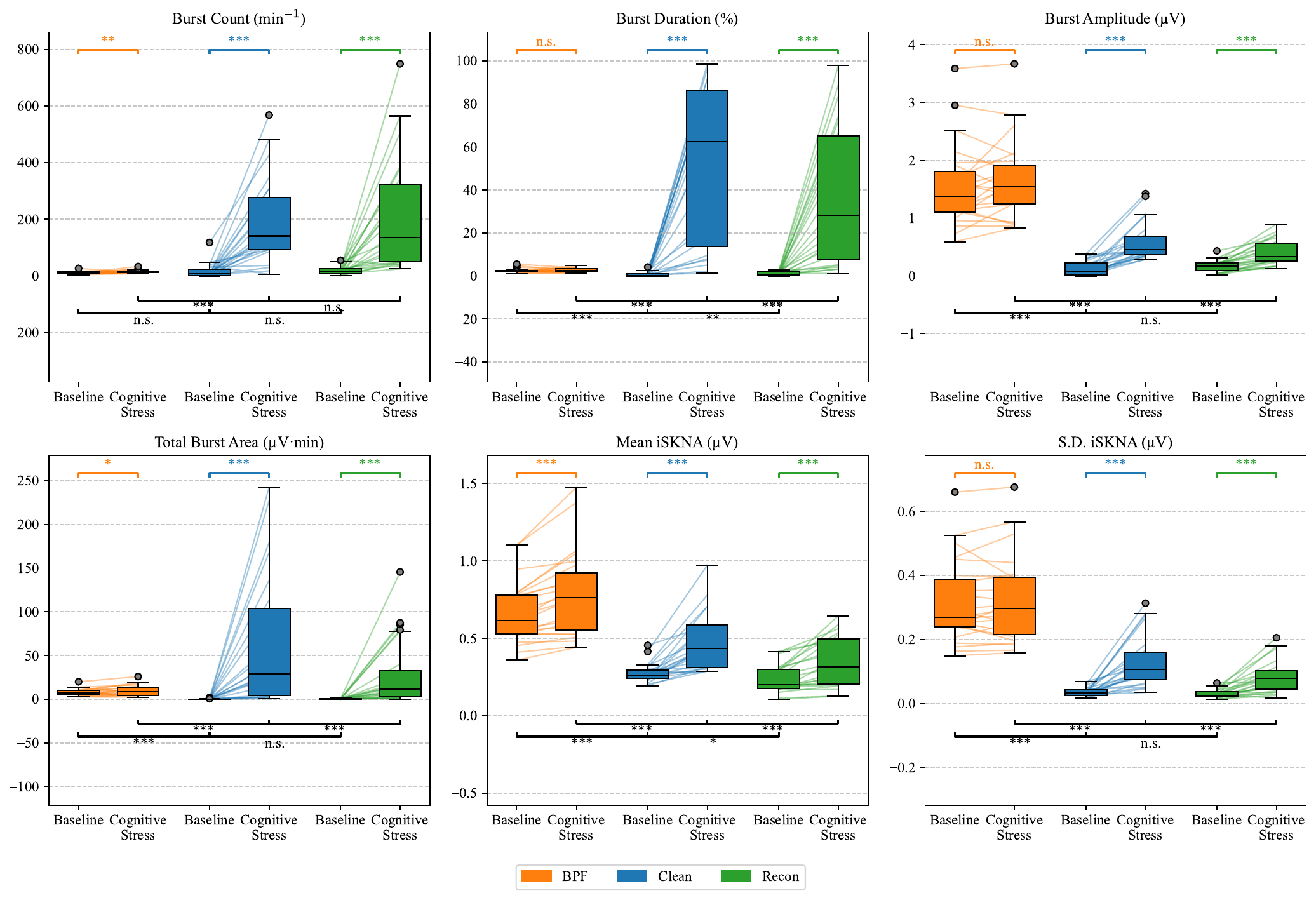}
    \caption{Boxplots of SKNA feature values for BPF (orange), reconstructed (green), and clean (blue) signals under simulated EMG contamination at SNR –4 dB, shown separately for baseline and SNS stimulation through Stroop cognitive task conditions. Statistical significance from Wilcoxon signed-rank tests is indicated above each comparison (*: p $\leq$ 0.05, **: p $\leq$ 0.01, ***:p $\leq$ 0.001, n.s. = not significant).}
    \label{fig:box-4}
\end{figure*}

\begin{figure*}
    \centering
    \includegraphics[width=0.98\linewidth]{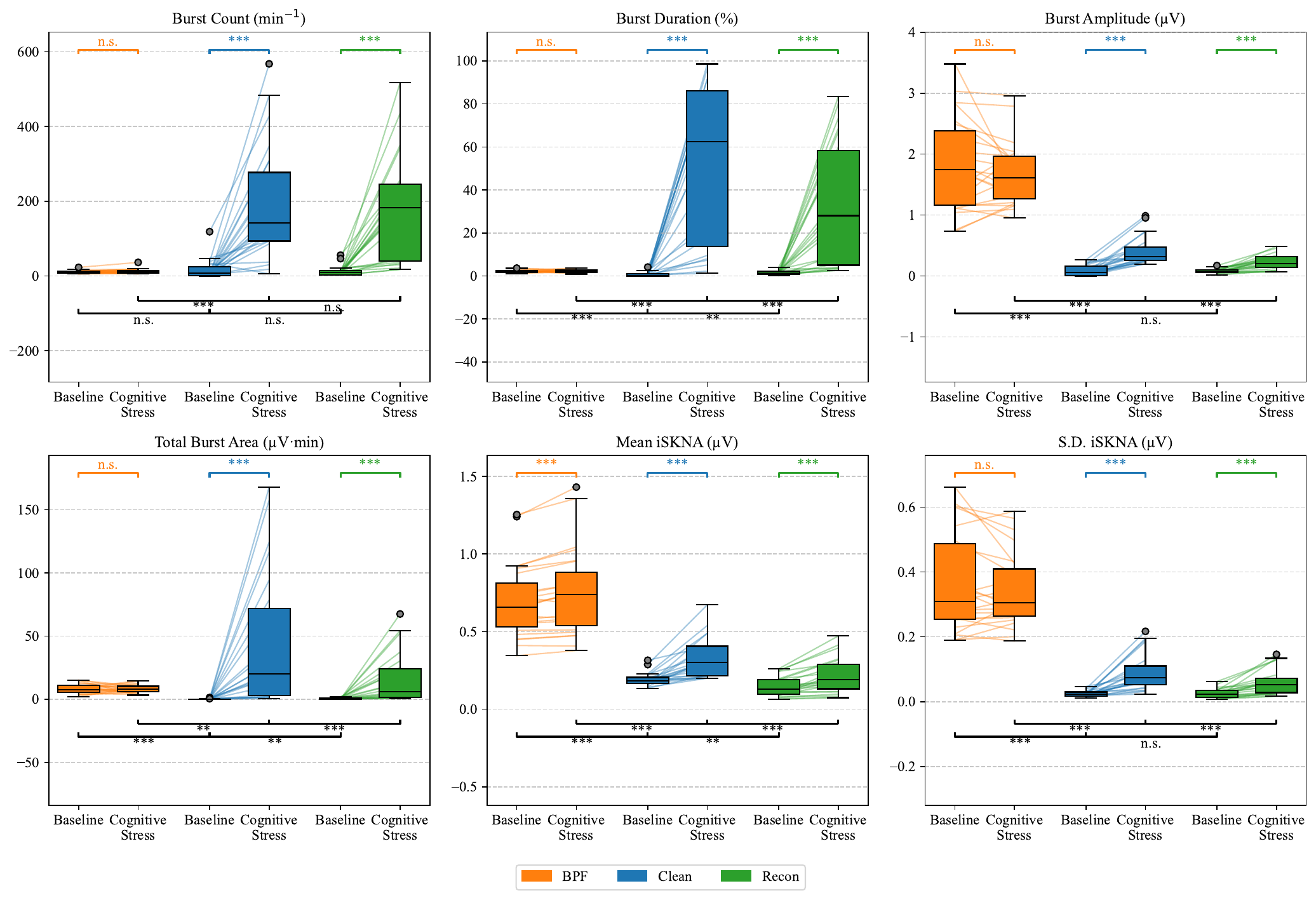}
    \caption{Boxplots of SKNA feature values for BPF (orange), reconstructed (green), and clean (blue) signals under simulated EMG contamination at SNR –8 dB, shown separately for baseline and SNS stimulation through Stroop cognitive task conditions. Statistical significance from Wilcoxon signed-rank tests is indicated above each comparison (*: p $\leq$ 0.05, **: p $\leq$ 0.01, ***:p $\leq$ 0.001, n.s. = not significant).}
    \label{fig:box-8}
\end{figure*}

\begin{table}[!t]
\caption{Statistical Analysis for iSKNA-derived features in discriminating between baseline and Cognitive Stress under simulated EMG contamination at SNR~\SI{-4}{\decibel}.
\label{tab:fisher-auroc-4dB}}
\centering
\begin{tabular}{|c|c|c|c|c|}
\hline Feature & Metric & BPF & Clean & Recon \\
\hline \multirow{2}{*}{Burst Count} 
 & Fisher’s Ratio & 0.2976 & 1.3523 & 1.0456 \\
 & AUROC          & 0.7335 & 0.9357 & 0.9622 \\
\hline \multirow{2}{*}{Burst Duration} 
 & Fisher’s Ratio & 0.0369 & 2.1213 & 1.3618 \\
 & AUROC          & 0.6200 & 0.9868 & 0.9811 \\
\hline \multirow{2}{*}{Burst Amplitude} 
 & Fisher’s Ratio & 0.0149 & 1.8424 & 1.1151 \\
 & AUROC          & 0.5690 & 0.9735 & 0.8601 \\
\hline \multirow{2}{*}{Burst Total Area} 
 & Fisher’s Ratio & 0.0683 & 0.7555 & 0.5544 \\
 & AUROC          & 0.5766 & 0.9905 & 0.9660 \\
\hline \multirow{2}{*}{Mean iSKNA} 
 & Fisher’s Ratio & 0.1306 & 1.0676 & 0.3476 \\
 & AUROC          & 0.6200 & 0.9093 & 0.7051 \\
\hline \multirow{2}{*}{S.D. iSKNA} 
 & Fisher’s Ratio & 0.0003 & 1.4110 & 1.0141 \\
 & AUROC          & 0.5123 & 0.9565 & 0.8620 \\
\hline
\end{tabular}
\end{table}

\begin{table}[!t]
\caption{Statistical Analysis for iSKNA-derived features in discriminating between baseline and Cognitive Stress under simulated EMG contamination at SNR~\SI{-8}{\decibel}.
\label{tab:fisher-auroc-8dB}}
\centering
\begin{tabular}{|c|c|c|c|c|}
\hline Feature & Metric & BPF & Clean & Recon \\
\hline \multirow{2}{*}{Burst Count} 
 & Fisher’s Ratio & 0.0174 & 1.3508 & 1.4846 \\
 & AUROC          & 0.5123 & 0.9357 & 0.9584 \\
\hline \multirow{2}{*}{Burst Duration} 
 & Fisher’s Ratio & 0.0124 & 2.1214 & 1.3122 \\
 & AUROC          & 0.5652 & 0.9868 & 0.9773 \\
\hline \multirow{2}{*}{Burst Amplitude} 
 & Fisher’s Ratio & 0.0213 & 1.8427 & 1.5551 \\
 & AUROC          & 0.4839 & 0.9735 & 0.8979 \\
\hline \multirow{2}{*}{Burst Total Area} 
 & Fisher’s Ratio & 0.0001 & 0.7557 & 0.6621 \\
 & AUROC          & 0.4991 & 0.9905 & 0.9357 \\
\hline \multirow{2}{*}{Mean iSKNA} 
 & Fisher’s Ratio & 0.0218 & 1.0686 & 0.3424 \\
 & AUROC          & 0.5652 & 0.9093 & 0.6938 \\
\hline \multirow{2}{*}{S.D. iSKNA} 
 & Fisher’s Ratio & 0.0123 & 1.4117 & 0.7184 \\
 & AUROC          & 0.4858 & 0.9565 & 0.8129 \\
\hline
\end{tabular}
\end{table}

Post-hoc analyses of SKNA-derived features revealed that reconstruction preserved sympathetic activity patterns more faithfully than BPF signals across both noise levels (Figures \ref{fig:box-4} and \ref{fig:box-8}). For SNR –4 dB (Table \ref{tab:fisher-auroc-4dB}), Fisher’s ratio values for reconstructed signals exceeded those of BPF signals in all features. AUROC values also increased markedly after reconstruction, reaching $\geq$ 0.96 for burst count, burst duration, and burst total area, approaching the performance of clean signals.

At SNR –8 dB, reconstruction continued to outperform BPF signals on all features (Table \ref{tab:fisher-auroc-8dB}), though absolute values were generally lower than at –4 dB. AUROC improvements were consistent, with several features (burst count, burst duration, burst total area) exceeding 0.93 after reconstruction, indicating preserved discriminability between baseline and cognitive stress conditions despite severe noise contamination.

Statistical comparisons indicated that BPF signals often failed to capture baseline–stimulation differences present in the clean data, whereas reconstructed signals restored many of these contrasts to statistically significant levels (Figures \ref{fig:box-4} and \ref{fig:box-8}). This preservation of condition-dependent changes suggests that the denoising process did not simply smooth away bursts or amplitude variations but retained physiologically relevant fluctuations. Overall, reconstruction improved both effect size (Fisher’s ratio) and classification-oriented separability (AUROC) across all features. This supports the feasibility of applying the proposed method in scenarios with substantial EMG contamination while maintaining sensitivity to sympathetic activation patterns.

\subsection{Classification Accuracy}
Classification analyses were performed to assess whether reconstructed signals preserved the ability to discriminate between baseline and cognitive stress conditions.

At SNR –4 dB (Table \ref{tab:classification-4dB}), BPF signals yielded modest classification performance, with LR achieving the highest accuracy (65.22\%) and AUC (0.75) among the three classifiers tested. Reconstruction markedly improved performance, raising RF accuracy to 97.83\% (AUC = 0.96), SVM to 86.96\% (AUC = 0.95), and Logistic Regression (LR) to 86.96\% (AUC = 0.98). These values approached those of clean signals, where RF reached 93.48\% accuracy and AUC = 0.97. Improvements were also reflected in sensitivity, specificity, and F1 score, with reconstructed RF achieving perfect specificity and near-perfect sensitivity.

At the more challenging SNR –8 dB level (Table \ref{tab:classification-8dB}), BPF signals exhibited poor classification, with RF, SVM, and LR accuracies near chance (50–56.52\%) and low AUCs (0.33–0.51). Reconstruction again restored classification performance to near-clean levels, with RF achieving 91.3\% accuracy (AUC = 0.97), SVM 86.96\% accuracy (AUC = 0.92), and LR 89.13\% accuracy (AUC = 0.94). These gains were consistent across sensitivity, specificity, and F1 score metrics.

\begin{figure*}
    \centering
    \includegraphics[width=0.98\linewidth]{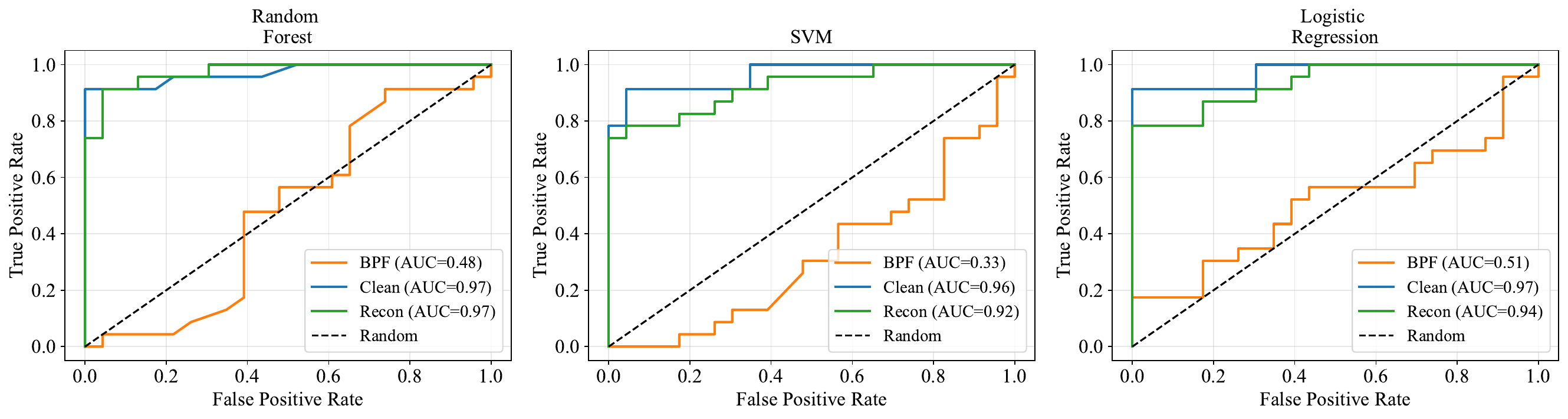}
    \caption{Receiver operating characteristic (ROC) curves for distinguishing baseline from cognitive stress conditions at SNR –8 dB using features derived from BPF, clean, and reconstructed iSKNA signals. Results are shown for Random Forest, Support Vector Machine (SVM), and Logistic Regression classifiers.}
    \label{fig:roc_curves}
\end{figure*}

\begin{figure}
    \centering
    \includegraphics[width=0.98\linewidth]{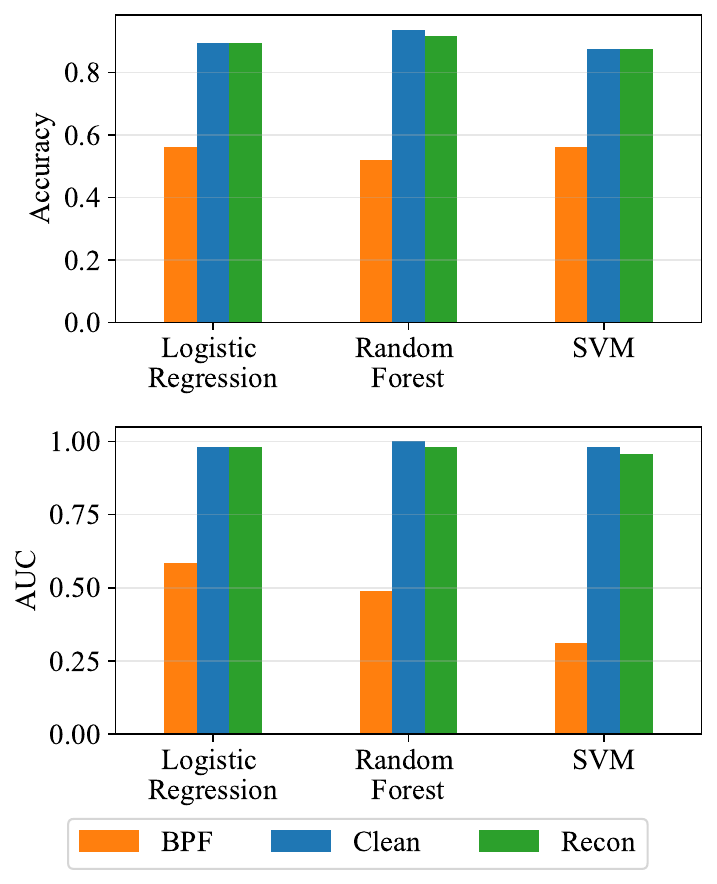}
    \caption{Cross-validation accuracy and area under the ROC curve (AUC) for baseline vs. cognitive stress classification at SNR –8 dB using features from BPF, clean, and reconstructed SKNA signals.}
    \label{fig:performance_bars}
\end{figure}

ROC curves for SNR –8 dB (Figure \ref{fig:roc_curves}) illustrate the poor separability of BPF signals and the substantial improvement achieved after reconstruction, with curves closely matching those from clean data. Similarly, cross-validation comparisons (Figure \ref{fig:performance_bars}) show that reconstruction recovers both accuracy and AUC to levels approaching clean performance across all classifiers.

Overall, these results indicate that the reconstruction method not only suppresses EMG contamination but also preserves discriminative features necessary for reliable condition classification, even under severe noise conditions.

\begin{table}[!t]
\caption{Classification performance for distinguishing baseline from cognitive stress at SNR~\SI{-4}{\decibel}.
\label{tab:classification-4dB}}
\centering
\begin{tabular}{|c|c|c|c|c|c|c|}
\hline Signal & Classifier & AUC & Acc. & Sens. & Spec. & F1 Score \\
\hline \multirow{3}{*}{BPF} 
 & RF  & 0.63 & 56.52 & 60.87 & 52.17 & 58.33 \\
 & SVM & 0.39 & 52.17 & 47.83 & 56.52 & 50.00 \\
 & LR  & 0.75 & 65.22 & 60.87 & 69.57 & 63.64 \\
\hline \multirow{3}{*}{Clean} 
 & RF  & 0.97 & 93.48 & 91.30 & 95.65 & 93.33 \\
 & SVM & 0.96 & 86.96 & 73.91 & 100.0 & 85.00 \\
 & LR  & 0.97 & 89.13 & 78.26 & 100.0 & 87.80 \\
\hline \multirow{3}{*}{Recon} 
 & RF  & 0.96 & 97.83 & 95.65 & 100.0 & 97.78 \\
 & SVM & 0.95 & 86.96 & 73.91 & 100.0 & 85.00 \\
 & LR  & 0.98 & 86.96 & 73.91 & 100.0 & 85.00 \\
\hline
\end{tabular}
\end{table}

\begin{table}[!t]
\caption{Classification performance for distinguishing baseline from cognitive stress at SNR~\SI{-8}{\decibel}.
\label{tab:classification-8dB}}
\centering
\begin{tabular}{|c|c|c|c|c|c|c|}
\hline Signal & Classifier & AUC & Acc. & Sens. & Spec. & F1 Score \\
\hline \multirow{3}{*}{BPF} 
 & RF  & 0.48 & 50.00 & 47.83 & 52.17 & 48.89 \\
 & SVM & 0.33 & 54.35 & 73.91 & 34.78 & 61.82 \\
 & LR  & 0.51 & 56.52 & 56.52 & 56.52 & 56.52 \\
\hline \multirow{3}{*}{Clean} 
 & RF  & 0.97 & 93.48 & 91.30 & 95.65 & 93.33 \\
 & SVM & 0.96 & 86.96 & 73.91 & 100.0 & 85.00 \\
 & LR  & 0.97 & 89.13 & 78.26 & 100.0 & 87.80 \\
\hline \multirow{3}{*}{Recon} 
 & RF  & 0.97 & 91.30 & 91.30 & 91.30 & 91.30 \\
 & SVM & 0.92 & 86.96 & 73.91 & 100.0 & 85.00 \\
 & LR  & 0.94 & 89.13 & 78.26 & 100.0 & 87.80 \\
\hline
\end{tabular}
\end{table}

\section{Conclusion}
We presented a deep learning–based approach for reconstructing SKNA from EMG-contaminated recordings, achieving substantial improvements in signal quality, feature preservation, and classification accuracy even at severe noise levels. Unlike traditional bandpass filtering, the method selectively removes overlapping EMG while retaining physiologically relevant sympathetic bursts, enabling more reliable SKNA monitoring in movement-rich environments.

\section{Significance}
Our previous conference paper was the first to explicitly identify EMG contamination as a critical limitation for SKNA measurement \cite{baghestani2024towards}. It applied deep learning to detect contaminated segments for removal, but this approach inevitably discards data and prevents analysis of useful features within those intervals. In contrast, the present study is the first to demonstrate a reconstruction framework that restores SKNA from EMG-dominated recordings rather than excluding them. This shift, from discarding to reconstructing, enables the preservation of physiologically relevant sympathetic bursts even during sustained muscle activity.

The significance lies in SKNA’s potential as a noninvasive biomarker of sympathetic activity. SKNA can be extracted from conventional ECG electrodes, making it feasible for continuous, long-term recording in both clinical and everyday settings. By providing a practical solution to the muscle artifact problem, our work lays the foundation for SKNA integration into portable and wearable technologies. Such devices could extend autonomic monitoring to real-world contexts, supporting applications in cardiovascular disease management, stress and fatigue detection, sleep assessment, and emotion recognition.

In summary, this study establishes a methodological advance that directly enhances the feasibility of SKNA as a noninvasive and real-time marker of sympathetic dynamics.

\section{Limitations and Future Work}
This study has some limitations that point to important directions for future work. First, we evaluated the proposed denoising protocol only under cognitive stress as a representative condition of sympathetic SNS activation. While useful for proof of concept, this restricted focus does not capture the diversity of physiological states in which SKNA may be modulated. Future studies should expand to include multiple forms of SNS activation and a larger pool of subjects, enabling stronger conclusions about generalizability.

Second, our framework relied on bandpass filtering to define the SKNA band prior to autoencoder-based denoising. Prior studies suggest that SNS-related information extends beyond the canonical frequency range. A more flexible approach that operates directly on broadband inputs, without presupposing a fixed frequency band, may better preserve physiologically relevant components of the signal.

Third, the noise model used in this study was based on additive EMG segments drawn from an independent dataset. While this provides controlled training conditions, it may not fully capture the complex spectral and temporal overlap between EMG and SKNA in real recordings. Future work will incorporate simultaneously collected EMG and ECG data, by recording two ECG channels, one stable and one subject to motion artifacts, to provide more realistic and subject-specific noise contamination.

By addressing these limitations, future studies can strengthen the evidence that the proposed autoencoder is not only an effective denoising tool in controlled conditions but also a generalizable method for real-world SKNA monitoring.

\bibliographystyle{IEEEtran}
\bibliography{bibliography}

\end{document}